\documentclass[letterpaper, 10 pt, conference]{ieeeconf} 
\IEEEoverridecommandlockouts                            
\usepackage{amsmath}
\usepackage{amsfonts}
\usepackage[linesnumbered,ruled]{algorithm2e}
\usepackage{array}
\usepackage{textcomp}
\usepackage{stfloats}
\usepackage{url}
\usepackage{verbatim}
\usepackage{subcaption}
\usepackage{graphicx}
\usepackage{cite}
\usepackage{multirow}
\usepackage{makecell}
\usepackage{setspace}
\usepackage{cellspace}
\usepackage{adjustbox}
\usepackage{booktabs} 
\usepackage{caption}
\usepackage{color,xcolor}
\definecolor{YoimiyaOrange}{RGB}{239,145,099}

\title{\LARGE \bf
DeepMF: Deep Motion Factorization for Closed-Loop Safety-Critical Driving Scenario Simulation
}

\author{Yizhe Li, Linrui Zhang, Xueqian Wang, Houde Liu, Bin Liang
\thanks{This work was supported by the Natural Science Foundation of China under Grant 92248304 and the Shenzhen Science Fund for Distinguished Young Scholars under Grant RCJC20210706091946001. \textit{(Corresponding author: Houde Liu)}}
\thanks{All authors are with the Center for Artificial Intelligence and Robotics, Shenzhen International Graduate School, Tsinghua University, Shenzhen 518055, China. \hspace{0pt}(e-mails: \hspace{0pt}li-yz23@mails.tsinghua.edu.cn; \hspace{0pt}zhanglr\hspace{0pt}.a\hspace{0pt}uto@gmail.com; \hspace{0pt}wang.xq@sz.tsinghua.edu.cn; \hspace{0pt}liu.hd@sz.\hspace{0pt}tsinghua\hspace{0pt}.edu\hspace{0pt}.c\hspace{0pt}n; \hspace{0pt}liangbin@mail.tsinghua.edu.cn)}
}

\begin{document}
\maketitle
\thispagestyle{empty}
\pagestyle{empty}


\begin{abstract}
Safety-critical traffic scenarios are of great practical relevance to evaluating the robustness of autonomous driving (AD) systems. Given that these long-tail events are extremely rare in real-world traffic data, there is a growing body of work dedicated to the automatic traffic scenario generation. However, nearly all existing algorithms for generating safety-critical scenarios rely on snippets of previously recorded traffic events, transforming normal traffic flow into accident-prone situations directly. In other words, safety-critical traffic scenario generation is hindsight and not applicable to newly encountered and open-ended traffic events.
In this paper, we propose the Deep Motion Factorization (DeepMF) framework, which extends static safety-critical driving scenario generation to closed-loop and interactive adversarial traffic simulation. DeepMF casts safety-critical traffic simulation as a Bayesian factorization that includes the assignment of hazardous traffic participants, the motion prediction of selected opponents, the reaction estimation of autonomous vehicle (AV) and the probability estimation of the accident occur. All the aforementioned terms are calculated using decoupled deep neural networks, with inputs limited to the current observation and historical states. Consequently, DeepMF can effectively and efficiently simulate safety-critical traffic scenarios at any triggered time and for any duration by maximizing the compounded posterior probability of traffic risk. Extensive experiments demonstrate that DeepMF excels in terms of risk management, flexibility, and diversity, showcasing outstanding performance in simulating a wide range of realistic, high-risk traffic scenarios.

\end{abstract}

\section{INTRODUCTION}
Ensuring the safety of autonomous driving is crucial, as it directly impacts public trust and acceptance of this technology \cite{ral1}. Before AVs can be fully deployed in the real world, it is essential to test the robustness of AD systems across a wide range of potential driving scenarios. Generally, there are two types of test data: one is collected from the real world, such as the Waymo \cite{waymo} and nuScenes \cite{nuscenes} datasets, and the other is generated by simulators, which create test scenarios by adjusting parameters related to driving safety, such as traffic flow\cite{metadrive}, vehicle behavior patterns\cite{KING}, and weather conditions\cite{weather}.

Scenarios collected from the real world most accurately reflect actual traffic conditions and natural human driving behaviors. In everyday driving situations, normal traffic scenes are predominant, while safety-critical scenarios—such as a neighboring vehicle suddenly cutting in or a leading vehicle making an emergency stop—are exceedingly rare. This discrepancy contributes to the long-tail distribution problem within the dataset. Nevertheless, these extreme scenarios are essential for evaluating the robustness of AD algorithms. ISO 21448 \cite{iso21448} also emphasizes the importance of continuous improvement in understanding and managing safety-critical scenarios.

On the other side, leveraging traffic simulators can automatically generate risk scenarios by adjusting parameters, such as increasing traffic density or programming surrounding vehicles (SVs) to behave unpredictably. However, this method requires careful manual adjustments of parameters and introduces a degree of randomness. Although the generation of safety-critical scenarios is virtually limitless with the aid of traffic simulators, the behavior of vehicles in these generated risk scenarios may not align with the intentions of human drivers, leading to a decrease in naturalness\cite{survey}. 

In our previous work \cite{cat}, we found that combining the two methods mentioned above, that is, converting a large number of normal scenes into safety-critical scenes in the simulator, can effectively test AD systems and conduct adversarial training. However, the previous work \cite{cat} still has many limitations, such as adversary selection relying on manual labels and risk scenarios being open-loop, which means it cannot respond to the reaction of AV in real time and is limited in generation time.
In this paper, we propose a light and closed-loop safety-critical scenario simulation framework called DeepMF, which considers both risk and realism. 
It is highly interactive which means it can respond to the continuously changing behavior of the AV in real-time and generate dynamic risk scenarios contrapuntally. DeepMF learns natural driving behaviors from real-world driving logs and aims to maximize the risk probability of the generated scenarios based on the deep-bayesian scenario factorization technique. The adversarial scenario generation problem is decomposed into the opponent forecasting and interdependent standard motion prediction sub-problems, which are solved step by step. 
At different time steps, DeepMF replans attack behaviors based on the newly observed traffic environment.
The decoupled deep neural network that implements aforementioned theoretical analysis leverages vectorized and rasterized information simultaneously to capture the complex features of agents based on the global scene.

Our contributions are as follows:

(i) We propose the closed-loop Deep Motion Factorization (DeepMF) framework, which factorizes the accident-prone scenarios simulation problem into four components: the hazardous evaluation of traffic participants, the motion prediction of selected opponents, the reaction estimation of ego vehicle and the probability estimation of accidents occur. This is all framed within the theoretical context of maximizing Bayesian posterior probability. Distinguished from previous methods for generating accident-prone events, DeepMF can effectively and efficiently simulate safety-critical driving scenarios at any triggered time and for any duration, without relying on the full replay of recorded traffic scenarios.

(ii) We design a heuristic task that generates supervisory signals autonomously from large-scale AD datasets, which can be used for training the feature extraction network. We also present an opponent prediction module that only requires a short segment of historical data to understand driving intentions of human drivers, which is able to make adversarial scores for all SVs and predict potential risk-causing agents among these automatically.

(iii) We conduct extensive experiments on over 1,500 traffic scenarios to demonstrate the efficiency of DeepMF. A total of 8 metrics are utilized to evaluate the algorithm's performance, such as algorithm runtime, the attack success rate, the naturalness of driving behaviors and agents' trajectories within the generated scenarios. Compared to state-of-the-art baselines, DeepMF is capable of generating highly risky, human-like, and diverse safety-critical scenarios in real-time more effectively.

\section{RELATED WORK}

\begin{figure*}[htbp]
    \centering
    \includegraphics[scale=0.55]{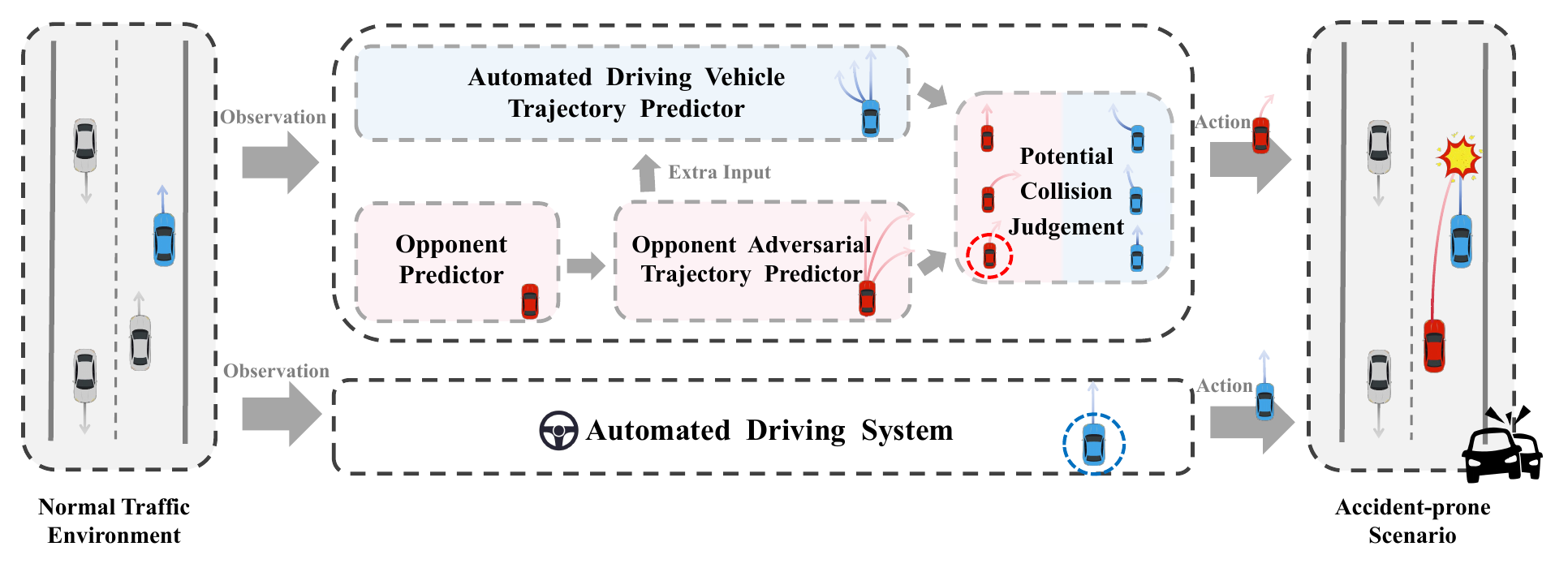}
    \caption{Overwiew of DeepMF. The opponent predictor selects the best attacker from surrounding agents. The trajectory predictor forecasts the trajectories of both the opponent and the ego vehicle, with the latter's prediction depending on the former's results. Next, a potential collision judgement is performed to choose the most aggressive yet reasonable opponent trajectory. It's important to note that the ego vehicle’s actual driving behavior is controlled by the independent planner.
    }
    \label{fig1}
\end{figure*}

Current relevant algorithms can be categorized into three primary approaches, knowledge-based generation, data-driven generation and adversarial-based generation \cite{survey, survey2}. 

Knowledge-based methods leverage predefined rules or constraint optimization to guide the generation process \cite{survey}. However, they rely on the completeness and accuracy of expert knowledge heavily and is hard to integrate with models. 
Data-driven methods builds density estimation models based on real-world data for scenario generation, yielding more natural scenarios \cite{survey, ral3, ral4}. Nonetheless, the scarcity of extreme scenarios in real-world data leads to insufficient relevant training data, resulting in lower accident-prone scenarios generation efficiency.  
Adversarial-based methods have the highest efficiency in generating challenging scenarios by actively attacking the AV. However, the generated scenarios may lack naturalness.

To leverage the complementary advantages of data-driven and adversarial-based methods, we propose a novel safety-critical scenario simulation framework.
The algorithm learns vehicle driving behavior from real-world scenarios and actively attacks the AV to generate high-adversarial traffic environment.

Recent works \cite{advsim, diffscene, art, cat, strive, KING} have explored the methods for generating adversarial scenarios based on deep-learning. 
ART \cite{art} perturbs normal trajectories with minor disturbances to maximize prediction errors, employing data augmentation and trajectory smoothing techniques to enhance the naturalness of predictions. However, it requires careful manual tuning of certain parameters, and the extra smoothing technique may effect the model's responsiveness to rapid changes. 
AdvSim \cite{advsim} generates adversarial trajectories by optimizing acceleration profiles based on a simplistic bicycle model, which may hard to capture the non-linear features in the complex traffic environment adequately.
CAT \cite{cat} introduces a closed-loop adversarial training framework, yet its selection of adversarial opponent relies on manual annotations within the waymo \cite{waymo} dataset, and it does not account for the dynamic interaction of the AV when crafting risky environment. STRIVE \cite{strive} models realistic traffic motion utilizing graph-based conditional variational autoencoder and optimizes in the latent space. However, it often takes several minutes to generate a scene because of the complex optimization steps.

Our approach is notably lightweight, capable of generating a safety-critical scenario within seconds. It only requires a short segment of vehicle history data to learn driving intentions and enables the selected the opponent vehicle (OV) to attack AV actively. By continuously adjusting the OV's trajectory in response to the behavioral changes of AV, the model can swiftly create risky and realistic scenarios.

\section{Problem Formulation}

\begin{figure*}[htbp]
    \centering
    \includegraphics[scale=0.52]{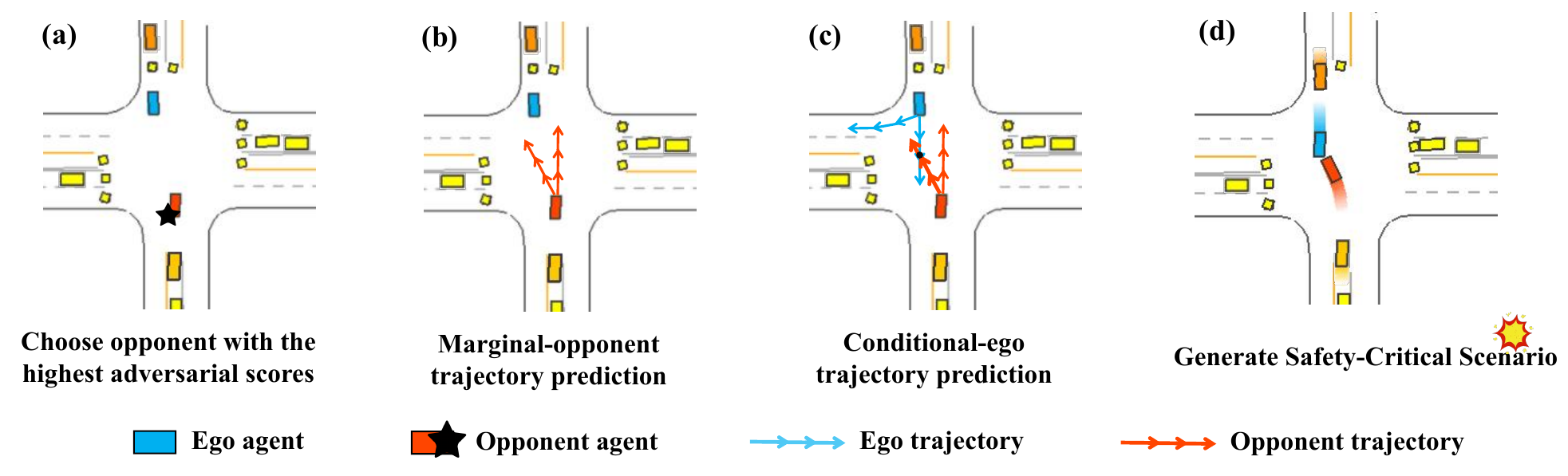}
    \caption{The process of DeepMF generating a challenging scenario begins with assigning adversarial scores to all surrounding vehicles, selecting the one with the highest score as the opponent to attack the ego vehicle. DeepMF then predicts the opponent's trajectories with corresponding scores based on the current environment, followed by predicting the possible reactive motions of the ego vehicle. Ultimately, it selects the opponent behavior most likely to cause an accident.}
    \label{fig2}
\end{figure*}

In the traffic scene, the observed states $X=(M, S)$ denotes the perception results of the traffic environment and surrounding agents. $M$ and $S$ denote the driving map and traffic participants respectively.
Based on the observed states $X$, we can predict the future states of agents $Y$.

We define a binary variable $Coll$, which takes the values $\{$$True, False$$\}$, indicating whether a collision event occurs between AV and SV in the scenario.
The expected value of a traffic accident occurring is denoted by $\mathbb{E}(Coll\!\mid\! X)$, presenting potential safety risk in the scenario.

\begin{align}
   & \mathbb{E}(Coll|X) 
    \!=\!1\!\times\!\mathbb{P}(Coll\!=\!True|X)\!+\!0\!\times\!\mathbb{P}(Coll\!=\!False|X) 
\end{align}

$\mathbb{P}(Coll\!=\!True|X)$ can be expressed as an integral over possible future states of the AV and OV:
\begin{align}
    \label{2}
    \int_{Y^{\!AV},Y^{\!OV}}\!\mathbb{P}(Coll\!=\!True, Y^{\!AV},Y^{\!OV}|X) 
\end{align}

Utilizing Bayesian formula, we can further decompose Eq. (\ref{2}) into a product of the joint distribution and the conditional probability:
\begin{align}
    &\!\int_{Y^{\!AV},Y^{\!OV}} \mathbb{P}(Y^{\!AV},Y^{\!OV}|X) \mathbb{P} (Coll\!=\!True|Y^{\!AV},Y^{\!OV},X) 
\end{align}

The joint distribution can be factorized as the prior $Y^{\!OV}$ and the conditional $Y^{\!AV}$ as follows:

\begin{align}
     \mathbb{P}(Y^{\!AV},Y^{\!OV}|X)
     =\mathbb{P}(Y^{\!OV}|X) \mathbb{P}(Y^{\!AV}|Y^{\!OV},X) 
\end{align}

Our objective is to maximize the possibility of a traffic collision occurring for AV, thereby generating accident-prone scene with the highest risk. From the above derivation, $\max \mathbb{E}(\text{Coll}\!\mid\!X)$ can be expressed as:
\begin{small}
\begin{align}
\label{factorization}
& \max_{Y^{\text{OV}}} \mathbb{P}(Y^{\text{OV}}\!\mid\!X) 
\int_{Y^{\text{AV}}} \mathbb{P}(Y^{\text{AV}}\!\mid\!Y^{\text{OV}}, X) 
\mathbb{P}(\text{Coll}\!=\!\text{True}\!\mid\!Y^{\text{AV}}, Y^{\text{OV}}, X) 
\end{align}
\end{small}

This deep-Bayesian scenario factorization formula in Eq. (\ref{factorization}) breaks down the safety-critical scenario generation problem into multiple subcomponents that can be efficiently solved, including opponent prediction, motion prediction, and possible collision detection.
The traffic agent with the highest safety risk for AV should be chosen as the opponent. Then based on the observation $X$, the motion is predicted, including behavior prediction for $Y^{\!OV}\!$ and interactive response prediction for $Y^{\!AV\!}$ conditioned on $Y^{\!OV}\!$. For each predicted pair of $Y^{\!OV\!}$ and $Y^{\!AV}$, the likelihood of a collision is evaluated.

Based on computations using this formula, the most adversarial behavior of the opponent $Y^{\!OV}$, which is most likely to lead to a traffic accident, is selected as the final prediction. The following sections will introduce the implementation of this idea.

\begin{algorithm}
\small
    \caption{Deep Motion Factorization for Safety-Critical Driving Scenario Simulation}
    \label{alg1} 
        \KwIn{Current traffic state $X$, AV policy/planner $\Pi$, Surrounding vehicles $\{SV_i\}_{i=1}^M$,  Adversarial score predictor $\phi$, Marginal trajectory predictor $\varphi_M$, Conditional trajectory predictor $\varphi_C$, Update cycle $T$}

        $\{s_i\}_{i=1}^M = \phi(X)$\tcp*{Predict adversarial score}
        $\text{IND} \sim Softmax\{s_1,\cdot\cdot\cdot,s_M\}$\tcp*{Sample OV index}
        $OV = SV_\text{IND}$\tcp*{Opponent Assignment}

        \For{$t \leftarrow 1$ \KwTo $\infty$}{
        \If{t \textbf{mod} T = 0}{
			$\left\{\left(Y_{j}^{OV}, P_{j}^{OV}\right)\right\}_{j=1}^{N_{1}} \sim \varphi_M(X)$\;
		\For{$j \leftarrow 1$ \KwTo $N_1$}{		
             $\left\{\left(Y_{k}^{AV}, P_{k}^{AV}\right)\right\}_{k=1}^{N_{2}} \sim \varphi_C(X,Y^{OV}_j)$\;
             \For{$k\leftarrow 1$ \KwTo $N_2$}{		
             ${ Coll }_{jk}=\operatorname{Intersect}\left(Y_{j}^{OV}, Y_{k}^{AV}\right)$\;
             }
             }
             ${Y}^{OV}=\underset{Y_{j}^{OV}}{\operatorname{argmax}} P\left(Y_{j}^{OV}\right) \sum P\left(Y_{k}^{AV}\right) Coll_{jk}$\;
  }
  \text{Simulator.step$(OV, {Y}^{OV})$}\tcp*{Maneuver OV}
  \text{Simulator.step$(AV, \Pi_{AV}(X))$}\tcp*{Maneuver AV}
  $X\leftarrow X_t$ \tcp*{Update traffic state}
        }
\end{algorithm}

\section{Method}
\subsection{Model Architecture Overview}
The architecture of our approach is shown in Fig. \ref{fig1}. 
The opponent prediction module aims to forecast the suitable attacking agent. It assigns each SV a predicted score, which reflects the possibility of being considered as the risky one. 
The trajectory prediction module forecasts the most aggressive path which is most likely to cause a safety-critical scenario for OV.
Possible trajectories of OV are predicted firstly. 
Based on the marginal-OV predicted results, conditional-AV behaviour which make possible response to newly predicted OV trajectories are also being forecasting. 
By iterating through all predicted OV-AV trajectory pairs, we evaluate their collision potential and choose the OV trajectory from the pair that causes the highest risk as the final prediction.
The implementation code of DeepMF is summarized in Alorithm \ref{alg1}. Fig. \ref{fig2} provides a clear explanation of this idea.

\subsection{Opponent Prediction}
\textit{1) Heuristic supervised signals generation Task.}
We expound on the simplistic design concept of this task below. The core idea is to generate supervisory signals for all SVs by determining whether they have a high level of interaction with the AV. Generally, SVs with similar driving routes and close proximity to the AV tend to interact more frequently with the AV, potentially having a greater impact on the AV's decision-making and being more suitable as attackers. In contrast, it is neither intuitive nor realistic for a distant SV to rapidly attack the AV by crossing the traffic flow in between. 

\begin{figure}[htbp]
    \centering
    \includegraphics[scale=0.3]{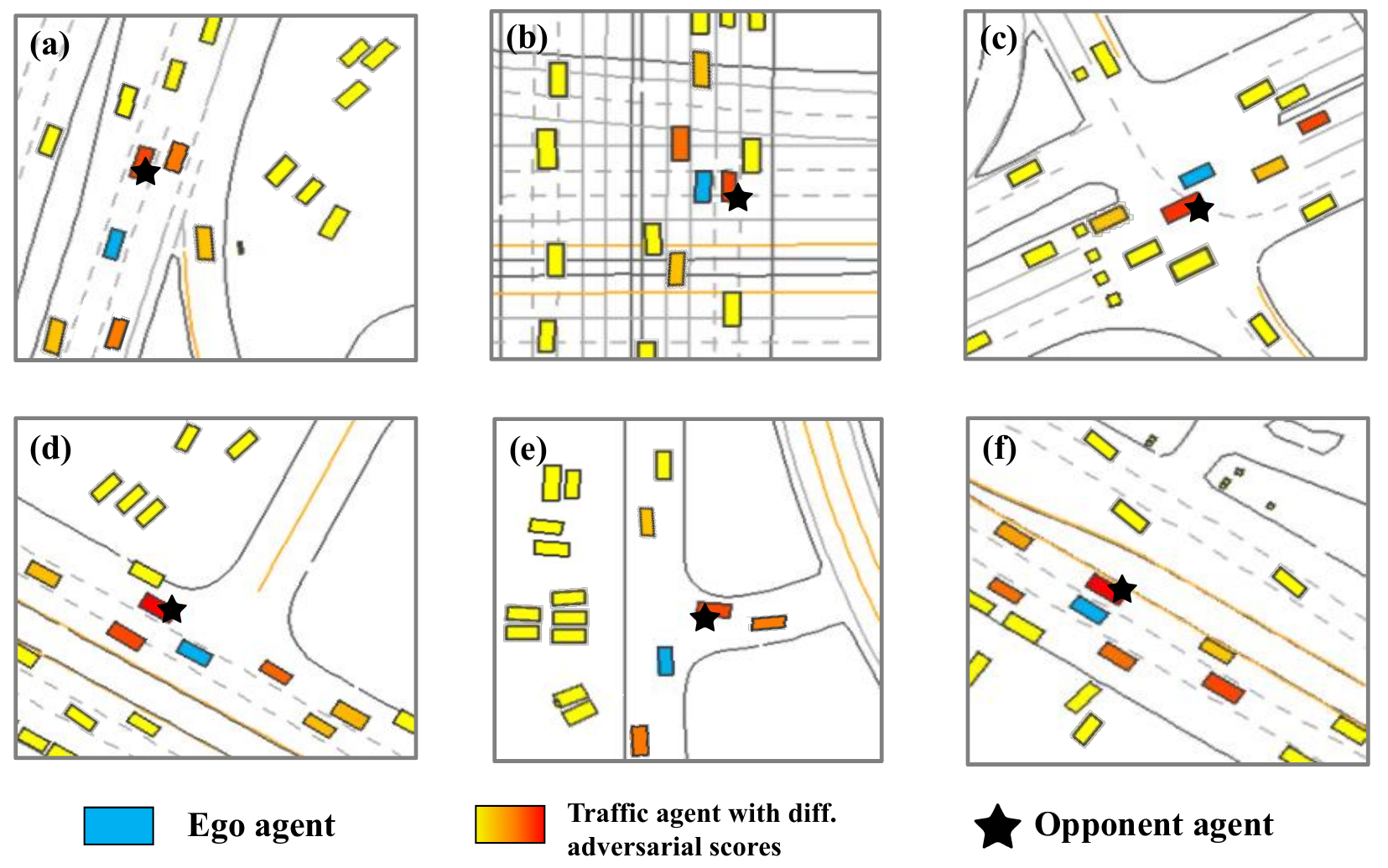}
    \caption{Output of the opponent prediction module on real-world scenarios. DeepMF assigns adversarial scores to all surrounding vehicles. The blue vehicle represents the ego. The adversarial scores for each vehicle are color-coded, ranging from yellow to red.}
    \label{fig3}
\end{figure}

To determine whether an SV has a high interaction possibility with the AV and is more likely to pose a potential safety risk, we calculate whether there is an overlap in the bounding boxes of their driving trajectories in the real-world dataset, which indicates whether their driving paths are similar. For example, situations where they drive sequentially on the same road. We also calculate whether the centroid distance between the two vehicles is less than the length of the AV, which indicates whether they are very close. For example, situations where they drive side by side in adjacent lanes.

This technique generates pseudo-labels, categorizing the SVs into positive samples with a high potential for causing safety risks and negative samples with lower potential. Then the opponent prediction network could learn the complex inherent structural information of the data based on these generated supervisory signals.

\textit{2) Opponent Prediction Network.}
The network utilizes both vectorized and rasterized information simultaneously. Vectorized features are extracted by the hierarchical graph neural model VectorNet \cite{vectornet}, which encodes the global map and traffic participants. The backbone network VGG16 \cite{vgg16} processes rasterized features where the traffic flow is represented as images. The prediction head consists of an MLP and a fully connected layer.

To address the issue of class imbalance between positive and negative samples, we applied focal loss as a mitigation strategy.
Compared to the cross-entropy loss function, it introduces a modulation factor that increases the model's focus on hard-to-classify samples and improves the model's performance in situations with class imbalance. The loss function is defined as:
\begin{equation}
L_{f l}=-\alpha_{\mathrm{t}}\left(1-p_{\mathrm{t}}\right)^\gamma \log \left(p_{\mathrm{t}}\right)
\end{equation}
where $\alpha_{\mathrm{t}}$ is used to balance the weights of positive and negative samples, while $\gamma$ adjusts the model's focus on samples of varying difficulty levels.

\begin{figure*}[htbp]
    \centering
    \includegraphics[scale=0.52]{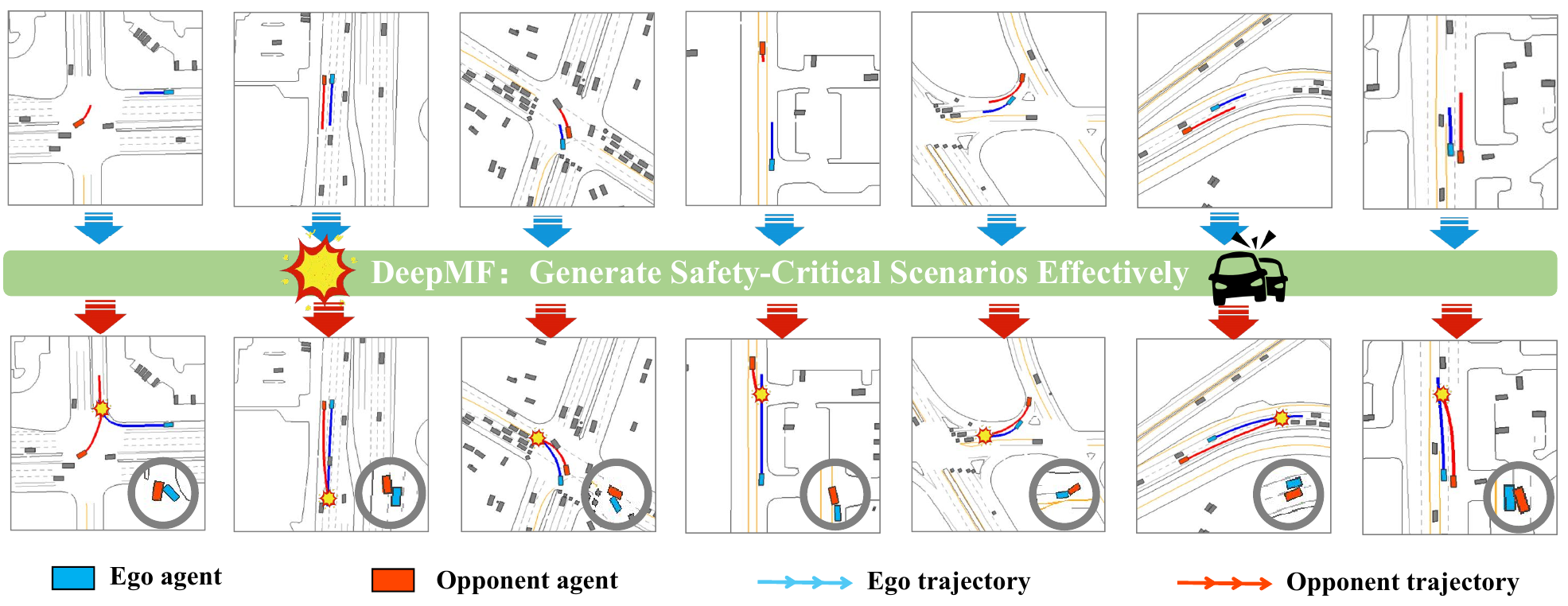}
    \caption{DeepMF can efficiently generate highly aggressive, natural, and diverse opponent motions across various traffic environments, producing safety-critical scenarios.}
    \label{fig4}
\end{figure*}

\textit{3) Implementation Detail.}
When generating supervisory signals and training the opponent prediction network, real-world driving scenarios are used, including environmental map and agents' trajectories within the episode. However, during inference, the network only requires the map information to understand the traffic scenario and a short-term historical trajectory of the vehicles to comprehend their driving behavior. During inference, the opponent prediction network assigns risk scores to all SVs and selects the one with the highest score as the active attacker to generate a safety-critical scenario.

Fig. \ref{fig3} shows output of the opponent prediction module on real datasets. The blue one is the AV. The adversarial scores predicted by this module for each vehicle are color-coded from yellow to red. The closer the color is to red, the higher the score, indicating a better fit for the role of an active attacker to create an accident-prone scenario. The agent marked with a star on the diagram is the one ultimately selected as the OV. For instance, in Fig. \ref{fig3}(a), the vehicle in front of the AV is predicted to be the best adversarial agent. If this OV suddenly decelerates, it will create a safety-critical environment, forcing the AV to make emergency maneuvers to navigate the scene successfully. Another example is shown in Fig. \ref{fig3}(b), where the predicted attacker is the SV driving in the lane adjacent to the AV. If this opponent suddenly attempts to merge into the AV's lane, a new adversarial scenario will be created.

\subsection{Opponent Trajectory Prediction}

\textit{1) Marginal-OV Trajectory Prediction.}
We utilize target-driven method DenseTNT \cite{densetnt} to predict possible OV trajectories with corresponding scores. The network extracts features of traffic flow using sparse representation and employs a dense goal encoder to generate probability distributions of goals to capture finer-grained information. 
The higher the relevant score of the predicted trajectory, the better the prediction result is, and the closer it aligns with the vehicle's actual driving behavior.

\textit{2) Conditional-AV Trajectory Prediction.}
We incorporate the predicted trajectory information of the OV from the aforementioned step as additional data into the vectorized and rasterized features, serving as supplementary guidance for AV trajectory prediction. Specifically, the predicted coordinates of the OV trajectories are added to the end of existing vectorized features and incorporated into the existing image with newly added channels.
It means that the AV trajectories, with their relevant predicted scores, are generated based on the real-time predicted trajectories of the OV from the above step. We also utilize DenseTNT as the conditional trajectory prediction model, where the input includes traffic context and the additional predicted information of the OV.

\begin{table*}[hb]
    \centering
    \caption{Comparison of Different Versions of DeepMF}
    \renewcommand{\arraystretch}{1.4}
    \label{tab1}
    \resizebox{\linewidth}{!}{%
    \begin{tabular}{@{\extracolsep{4pt}}lcccccccc@{}}
        \hline
        \multirow{2}{*}{Methods}
        & \multicolumn{3}{c}{Attack Results} & \multicolumn{2}{c}{Action Similarity} & \multicolumn{2}{c}{Trajectory Similarity} & \multirow{2}{*}{Generate Time (s)} \\
        \cline{2-4}
        \cline{5-6}
        \cline{7-8}
        &
        \begin{tabular}{c}
            Coll Rate (\%) 
        \end{tabular}
        & 
        \begin{tabular}{c}
            Coll Time (s)
        \end{tabular}
        & 
        \begin{tabular}{c}
            Coll Vel (m/s)   
        \end{tabular}
        & 
        \begin{tabular}{c}
            KL  
        \end{tabular}
        & 
        \begin{tabular}{c}
            Wasserstein  
        \end{tabular}
        & 
        \begin{tabular}{c}
            SSP   
        \end{tabular}
        & 
        \begin{tabular}{c}
            Hausdorff    
        \end{tabular}  
        \\
        \hline
        \textbf{Replay Planner} \\
        \hline
        DeepMF-G & $88$ & $\mathbf{4.11}$ & $\mathbf{5.70}$ & $1.30$ & $1.97$ & $1.39$ & $11.96$ & $\mathbf{1.39}$  \\
        DeepMF-S4 & $87$ & $4.12$ & $5.42$ & $1.26$ & $1.17$ & $1.41$ & $12.08$ & $1.69$  \\
        DeepMF-S2 & $91$ & $4.14$ & $4.98$ & $1.21$ & $1.06$ & $1.36$ & $11.98$ & $2.29$  \\
        DeepMF-S1 & $\mathbf{93}$ & $4.30$ & $4.22$ & $\mathbf{1.14}$ & $\mathbf{1.02}$ & $\mathbf{1.27}$ & $\mathbf{11.93}$ & $3.58$  \\ 
        \hline
        \textbf{IDM Planner} \\
        \hline
        DeepMF-G & $84$ & $\mathbf{4.06}$ & $\mathbf{5.51}$ & $1.16$ & $1.24$ & $2.59$ & $16.21$ & $\mathbf{1.41}$  \\
        DeepMF-S4 & $87$ & $4.10$ & $5.17$ & $1.44$ & $1.21$ & $2.30$ & $14.16$ & $1.73$ \\
        DeepMF-S2 & $86$ & $4.16$ & $4.80$ & $1.13$ & $1.11$ & $1.98$ & $\mathbf{12.61}$ & $2.38$  \\
        DeepMF-S1 & $\mathbf{88}$ & $4.24$ & $4.26$ & $\mathbf{1.12}$ & $\mathbf{1.10}$ & $\mathbf{1.64}$ & $12.85$ & $3.61$  \\ 
        \hline
    \end{tabular}  
    }
    \vspace{10pt}
\end{table*}

\textit{3) Potential Collision Judgement.}
We pair these predicted AV-OV trajectories and calculate the product of their respective predicted scores.
Additionally, we judge whether they would crash in that predicted situation. If a collision is predicted, we also record the expected time of the crash. 
A collision is considered to occur if the bounding boxes of the AV and OV overlap at the same time.

\textit{4) Opponent Trajectory Selection.}
If some AV-OV path pairs have potential collision, just within this range, we select the OV trajectory from the pair with highest multiplied score as final result.
Otherwise, we directly retrieval all pairs to seek it, who from the highest multiplied score pair. 

\section{EXPERIMENTS}
\subsection{Experimental Setup} 
\textit{1) Real-world Human Driving Dateset.} Waymo Open Motion Dataset (WODM), a driving dataset including abundant real-world traffic environment \cite{waymo}. We conduct experiments on 1,500 scenarios, with each episode lasting 9 seconds, using 1 second of historical motion data to predict the following 8 seconds of future trajectory.

\textit{2) Traffic Environment Simulation Platform.} We import experimental scenarios into the lightweight simulator MetaDrive \cite{metadrive}. 
Two automatic driving planners are utilized during the evaluation, which are the replay planner and the IDM planner. The replay planner directly paybacks 8 seconds future motion of AV from WODM driving log. The IDM planner replans the AV's behavior based on the newest traffic environment at each step, making it more interactive.

\textit{3) Adversarial scenario generation Baselines.} Five SOTA safety-critical scenario generation algorithms are considered as baselines, which are STRIVE, ART, BBO, BGA, and BRS. 
The latter three are bicycle-based models that are implemented based on the principles of the AdvSim \cite{advsim} framework. They utilize three different search algorithms which are Bayesian Optimization \cite{bbo}, Genetic Algorithm \cite{bga}, and Random Search \cite{brs} respectively. They optimize kinematic bicycle parameters based on black-box methods to generate safety-critical scenarios.

\begin{figure*}[ht]
    \centering
    \includegraphics[scale=0.53]{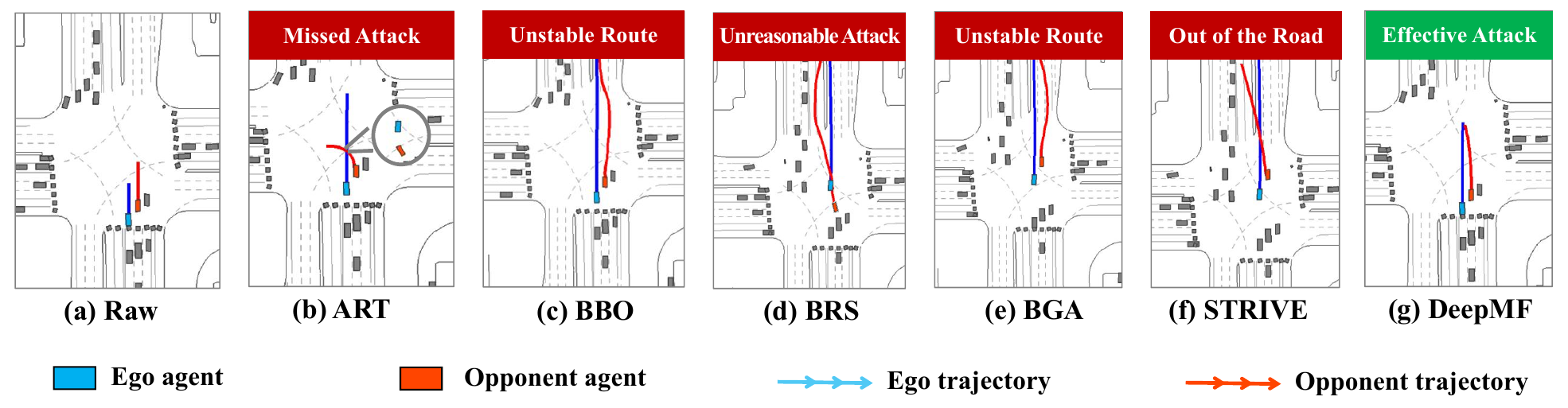}
    \caption{The qualitative evaluation results where various algorithms generate adversarial trajectories based on the same raw scenario. The blue vehicle is ego with future trajectory. The red one is opponent with future trajectory.}
    \label{fig6}
\end{figure*}

\subsection{Evaluation Metrics} 
We evaluate algorithms from two aspects, which are the efficiency of generating adversarial environments and the naturalness of the generated attack trajectories. 
The efficiency measurement mainly includes four metrics, which are collision rate, collision time, collision relative velocity and the consumption time of generating per scenario. 
The naturalness measurement mainly includes two aspects. On one hand, it measures the similarity of action distribution using Kullback–Leibler Divergence and Wasserstein Distance metrics. On the other hand, it assesses the similarity of the generated trajectories utilizing Symmetric Segment-Path Distance and Hausdorff Distance metrics.

\textit{1) Efficiency-level Evaluation.}
The collision rate indicates the proportion of accidents that actually occur in the generated scenarios. 
A higher collision rate suggests that more challenging scenarios are being created. 
In the event of an accident, collision time and collision relative speed of AV and OV are also recorded. 
Earlier collision time and higher collision relative velocity indicate higher risk and more severe accidents. Generation Time is used to assess the speed of the model's operation.

\textit{2) Naturalness-level Evaluation.}
It is used to assess the realism of the generated adversarial scenarios, specifically examining whether the driving behavior of the opponent vehicle resembles the decisions made by human drivers. Vehicle acceleration similarity is measured using Kullback–Leibler Divergence and Wasserstein Distance, while trajectory similarity is assessed through Symmetric Segment-Path Distance and Hausdorff Distance.

\subsection{Analysis of DeepMF Simulation.}
We conduct multiply experiments on four versions of the DeepMF algorithm, namely DeepMF-G, DeepMF-S1, DeepMF-S2, and DeepMF-S4. 
The G version operates as an open-loop method, generating 8 seconds of future adversarial trajectory all at once.
In contrast, the S version is a closed-loop method, where DeepMF replans the attack motion based on the currently observed traffic environment at different time steps. The S1 version refers to DeepMF replanning every second, while the S2 and S4 versions follow the same logic, replanning every 2 seconds and 4 seconds, respectively. 
We carried out two sets of experiments, one for replay planner and another for IDM planner. The experimental results are presented in Tab. \ref{tab1}.

The results show that DeepMF-S1 achieves the highest collision rate, reaching 93\% with the Replay planner and 88\% with the IDM planner. This may be because DeepMF-S1 is capable of making new and more aggressive decisions based on the most recently observed scene. 
Additionally, DeepMF-S1 excels in both action similarity and trajectory similarity measurements, indicating that its driving actions align more closely with the intentions of human drivers, resulting in more natural attack trajectories.
DeepMF-G achieves the shortest collision time since the start of challenging scenario generation and the highest relative speed of AV and OV when accident occur, suggesting that it could produce relatively more serious accident scenarios.

DeepMF-G achieves the shortest algorithm running time, taking 1.39 seconds with the Replay planner and 1.41 seconds with the IDM planner. 
It maybe because DeepMF-G generates the entire attack trajectory in one go, while the S versions need to interact with the environment and continuously replan based on the current status of the traffic scene, resulting in slightly longer processing times compared to the G version.

\begin{figure}[htbp]
    \centering
    \includegraphics[scale=0.28]{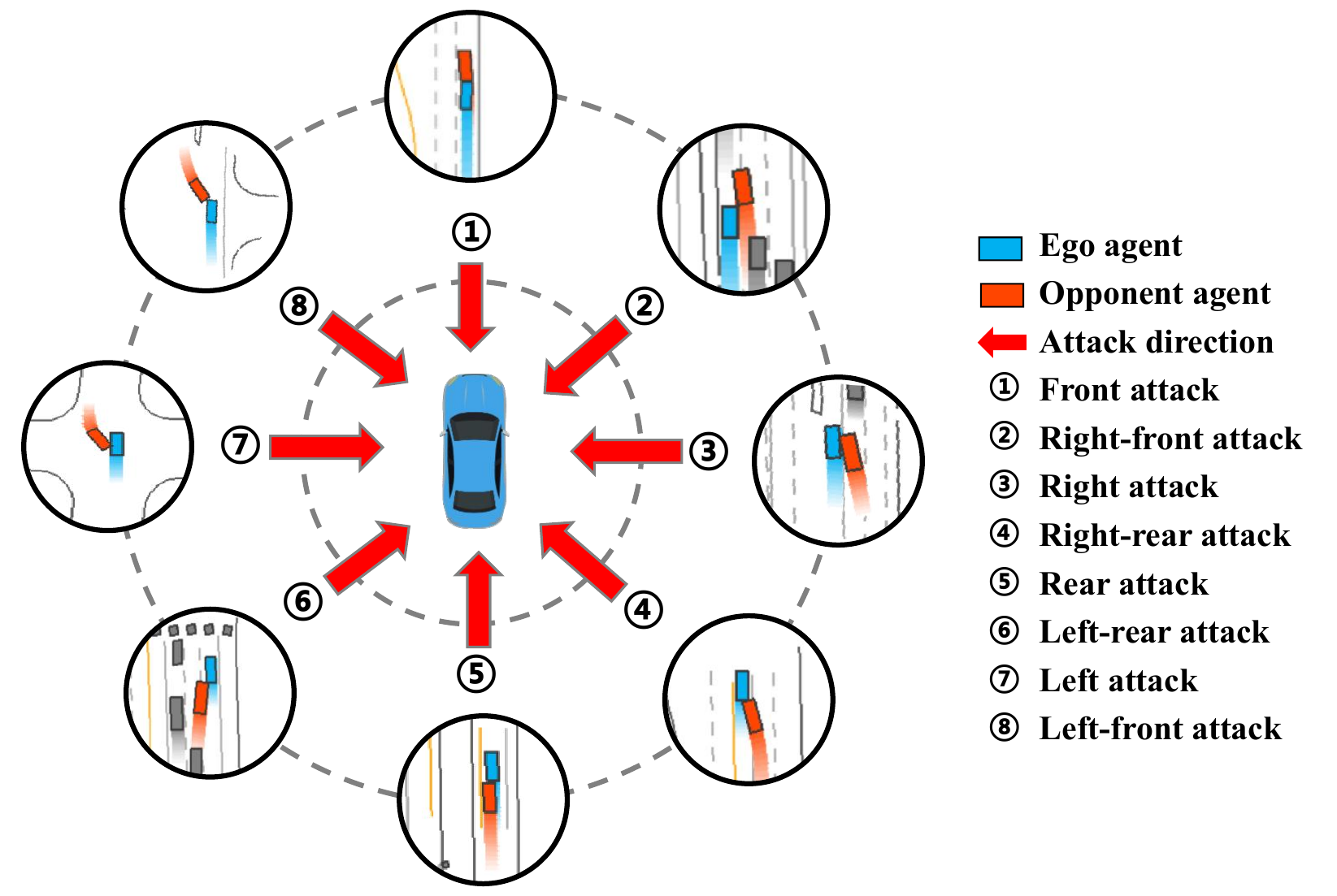}
    \caption{Eight types of accidents in safety-critical scenarios generated by DeepMF, showcasing attacks originating from various directions.}
    \label{fig5}
\end{figure}

Figure \ref{fig4} presents the qualitative evaluation results of DeepMF, generating accident-prone scenarios based on various traffic conditions.
Fig. \ref{fig5} visually illustrates eight different types of accidents generated by DeepMF, featuring attacks originating from various directions, including front attack, right-front attack, right attack, right-rear attack, rear attack, left-rear attack, left attack, and left-front attack.

\begin{table}[htbp] 
    \centering
    \caption{Efficiency-level Evaluation of Baselines}  
    \label{tab2}
    \large
    \renewcommand{\arraystretch}{1.6}  
    \resizebox{\linewidth}{!}{     
    \begin{tabular}{@{\extracolsep{4pt}}lcccc@{}}
        \hline
        Methods 
        &
        {
            \renewcommand{\arraystretch}{1}
            \begin{tabular}{c}
                Coll Rate \\
                $\left( \% \right)$
            \end{tabular}
        } 
        & 
        {
            \renewcommand{\arraystretch}{1}
            \begin{tabular}{c}
                Coll Time \\
                $\left( \mathit{s} \right)$
            \end{tabular}
        } 
        & 
        {
            \renewcommand{\arraystretch}{1}
            \begin{tabular}{c}
                Coll Vel \\
                $\left( \mathit{m/s} \right)$ 
            \end{tabular}
        } 
        &
        {
            \renewcommand{\arraystretch}{1}
            \begin{tabular}{c}
             Generate Time \\
             $\left( \mathit{s} \right)$
            \end{tabular}
        } \\
        \hline
        \textbf{Replay Planner} \\
        \Xhline{3\arrayrulewidth}
        STRIVE & $86$ & $4.65$ & $3.17$ & $187.61$ \\
        ART & $85$ & $4.38$ & $4.16$ & $64.94$ \\
        BBO & $83$ & $5.75$ & $3.87$ & $49.90$ \\
        BRS & $46$ & $4.92$ & $4.16$ & $\mathbf{1.55}$ \\ 
        BGA & $81$ & $6.02$ & $3.62$ & $44.32$ \\
        DeepMF & $\mathbf{93}$ & $\mathbf{4.30}$ & $\mathbf{4.22}$ & $3.58$ \\ 
        \hline
        \textbf{IDM Planner} \\
        \Xhline{3\arrayrulewidth}
        STRIVE & $78$ & $4.41$ & $3.85$ & $189.36$ \\
        ART & $72$ & $4.72$ & $4.18$ & $65.13$ \\
        BBO & $70$ & $5.34$ & $3.65$ & $41.65$ \\
        BRS & $34$ & $4.28$ & $2.92$ & $\mathbf{1.49}$ \\ 
        BGA & $74$ & $5.71$ & $3.93$ & $45.73$ \\
        DeepMF & $\mathbf{88}$ & $\mathbf{4.24}$ & $\mathbf{4.26}$ & $3.61$ \\ 
        \hline
    \end{tabular}  
    }
    \vspace{10pt}  
\end{table}

\subsection{Comparision of Baselines.} 
Tab. \ref{2} shows the efficiency-level evaluation results of baselines with Replay planner and IDM planner respectively. It shows that DeepMF achieves the highest collision rate, the shortest collision time, and the highest collision relative velocity, demonstrating its ability to generate more frequent and severe accidents.

In terms of average challenging scene generation time, BRS has the shortest duration, taking 1.55 seconds and 1.49 seconds with Replay planner and IDM planner, respectively. 
However, it also exhibits the lowest collision rates, at 46\% and 34\%, indicating its relatively poor aggressiveness. 
DeepMF ranks second for the scene generation time, with durations of 3.58 seconds and 3.61 seconds, significantly lower than the four algorithms STRIVE \cite{strive}, ART \cite{art}, BRS, and BGA, which all exceed 40 seconds. 
It demonstrates that DeepMF strikes a balance between aggressive and lightweight, attaining optimal attack outcome with a faster generation speed.
Figure \ref{fig6} shows the qualitative evaluation results, where various algorithms generate adversarial trajectories based on the same raw scenario. In the raw scenario, both the ego agent and the opponent agent are moving straight in the same direction. ART's sluggish motion attack results in no collision where Ego vehicle driving in front of Opponent vehicle. BBO and BGA attack successfully, but their generated trajectories are unstable. BRS exhibits excessive randomness, attacking from the opposite side of the lane, which deviates from normal driving behavior. STRIVE's attack fails and even exceeds the driving area directly. DeepMF controls opponent to gradually approach the AV, generating a natural trajectory and successfully executing the attack.

\begin{table}[htbp] 
    \centering
    \caption{Naturalness-level Evaluation of Baselines.}  
    \label{tab3}
    \renewcommand{\arraystretch}{1.4}  
    \resizebox{\linewidth}{!}{     
    \begin{tabular}{@{\extracolsep{4pt}}lcccc@{}}
        \hline
        
        \multirow{2}{*}{Methods} 
        & 
         \multicolumn{2}{c}{Action Similarity} & \multicolumn{2}{c}{Trajectory Similarity}  \\
            \cline{2-3}
            \cline{4-5}
        & 
        {
            \renewcommand{\arraystretch}{1}
            \begin{tabular}{c}
                KL 
            \end{tabular}
        } 
        & 

        {
            \renewcommand{\arraystretch}{1}
            \begin{tabular}{c}
                Wasserstein  
            \end{tabular}
        } 
        & 
        {
            \renewcommand{\arraystretch}{1}
            \begin{tabular}{c}
               SSP   
            \end{tabular}
        } 
        & 
        {
            \renewcommand{\arraystretch}{1}
            \begin{tabular}{c}
                Hausdorff   
            \end{tabular}
        }  
        \\
        \hline
        \textbf{Replay Planner} \\
        \Xhline{3\arrayrulewidth}
        STRIVE & $8.79$ & $8.77$ & $6.13$ & $94.85$   \\
        ART & $11.73$ & $15.27$ & $23.61$ & $38.64$   \\
        BBO & $1.19$ & $3.06$ & $2.64$ & $15.09$   \\
        BRS & $1.32$ & $4.32$ & $4.47$ & $20.02$  \\ 
        BGA & $2.26$ & $2.92$ & $2.51$ & $13.99$  \\
        DeepMF & $\mathbf{1.14}$ & $\mathbf{1.02}$ & $\mathbf{1.27}$ & $\mathbf{11.93}$  \\
        \hline
        \textbf{IDM Planner} \\
        \Xhline{3\arrayrulewidth}
        STRIVE & $6.52$ & $7.36$ & $6.89$ & $95.73$   \\
        ART & $14.72$ & $11.30$ & $24.87$ & $29.94$   \\
        BBO & $1.21$ & $3.29$ & $6.30$ & $14.18$   \\
        BRS & $1.26$ & $4.34$ & $4.69$ & $20.40$   \\ 
        BGA & $1.35$ & $3.35$ & $2.57$ & $16.82$   \\ 
        DeepMF & $\mathbf{1.12}$ & $\mathbf{1.10}$ & $\mathbf{1.64}$ & $\mathbf{12.85}$   \\ 
        \hline
    \end{tabular}  
    }
    \vspace{10pt}  
\end{table}

Tab. \ref{tab3} also presents the naturalness-level evaluation results of baselines with Replay planner and IDM planner respectively. 
It shows that DeepMF scores the lowest for both Kullback–Leibler Divergence and Wasserstein Distance, indicating that it achieves the highest action similarity, and its action distribution closely aligns with realistic scenarios. 
Additionally, DeepMF achieves the lowest scores in Symmetric Segment-Path Distance and Hausdorff Distance, signifying that the generated trajectories closely resemble the driving paths of human drivers.

\section{CONCLUSION}
In this paper, we propose the DeepMF framework, which is based on the deep Bayesian scenario factorization technique, designed to generate accident-prone scenarios. The framework breaks down the complex task of scenario generation into four key components, that is, the adversarial evaluation of traffic participants, the marginal-motion prediction of selected opponents, the conditional-reaction estimation of AV and the probability estimation of the collision happend.At different time steps, DeepMF replans attack behaviors based on the newly observed traffic environment. Experimental results show that DeepMF outperforms other baselines in efficiency-level and naturalness-level evaluations and is capable of generating diverse and challenging environments in a short amount of time.

\addtolength{\textheight}{-12cm} 

\bibliographystyle{IEEEtran}
\bibliography{reference}{}

\begin{thebibliography}{10}
\providecommand{\url}[1]{#1}
\csname url@rmstyle\endcsname
\providecommand{\newblock}{\relax}
\providecommand{\bibinfo}[2]{#2}
\providecommand\BIBentrySTDinterwordspacing{\spaceskip=0pt\relax}
\providecommand\BIBentryALTinterwordstretchfactor{4}
\providecommand\BIBentryALTinterwordspacing{\spaceskip=\fontdimen2\font plus
\BIBentryALTinterwordstretchfactor\fontdimen3\font minus \fontdimen4\font\relax}
\providecommand\BIBforeignlanguage[2]{{%
\expandafter\ifx\csname l@#1\endcsname\relax
\typeout{** WARNING: IEEEtran.bst: No hyphenation pattern has been}%
\typeout{** loaded for the language `#1'. Using the pattern for}%
\typeout{** the default language instead.}%
\else
\language=\csname l@#1\endcsname
\fi
#2}}

\bibitem{ral1}
H.~Woo, Y.~Ji, H.~Kono, Y.~Tamura, Y.~Kuroda, T.~Sugano, Y.~Yamamoto, A.~Yamashita, and H.~Asama, ``Lane-change detection based on vehicle-trajectory prediction,'' \emph{IEEE Robotics and Automation Letters}, vol.~2, no.~2, pp. 1109--1116, 2017.

\bibitem{waymo}
\BIBentryALTinterwordspacing
S.~Ettinger, S.~Cheng, B.~Caine, C.~Liu, H.~Zhao, S.~Pradhan, Y.~Chai, B.~Sapp, C.~Qi, Y.~Zhou, Z.~Yang, A.~Chouard, P.~Sun, J.~Ngiam, V.~Vasudevan, A.~McCauley, J.~Shlens, and D.~Anguelov, ``\BIBforeignlanguage{en-US}{Large scale interactive motion forecasting for autonomous driving: The waymo open motion dataset},'' in \emph{\BIBforeignlanguage{en-US}{2021 IEEE/CVF International Conference on Computer Vision (ICCV)}}, Oct 2021. [Online]. Available: \url{http://dx.doi.org/10.1109/iccv48922.2021.00957}
\BIBentrySTDinterwordspacing

\bibitem{nuscenes}
\BIBentryALTinterwordspacing
H.~Caesar, V.~Bankiti, A.~H. Lang, S.~Vora, V.~E. Liong, Q.~Xu, A.~Krishnan, Y.~Pan, G.~Baldan, and O.~Beijbom, ``\BIBforeignlanguage{en-US}{nuscenes: A multimodal dataset for autonomous driving},'' in \emph{\BIBforeignlanguage{en-US}{2020 IEEE/CVF Conference on Computer Vision and Pattern Recognition (CVPR)}}, Jun 2020. [Online]. Available: \url{http://dx.doi.org/10.1109/cvpr42600.2020.01164}
\BIBentrySTDinterwordspacing

\bibitem{metadrive}
\BIBentryALTinterwordspacing
Q.~Li, Z.~Peng, L.~Feng, Q.~Zhang, Z.~Xue, and B.~Zhou, ``\BIBforeignlanguage{en-US}{Metadrive: Composing diverse driving scenarios for generalizable reinforcement learning},'' \emph{\BIBforeignlanguage{en-US}{IEEE Transactions on Pattern Analysis and Machine Intelligence}}, p. 1–14, Jan 2022. [Online]. Available: \url{http://dx.doi.org/10.1109/tpami.2022.3190471}
\BIBentrySTDinterwordspacing

\bibitem{KING}
N.~Hanselmann, K.~Renz, K.~Chitta, A.~Bhattacharyya, and A.~Geiger, ``King: Generating safety-critical driving scenarios for robust imitation via kinematics gradients,'' in \emph{European Conference on Computer Vision}.\hskip 1em plus 0.5em minus 0.4em\relax Springer, 2022, pp. 335--352.

\bibitem{weather}
N.~Ruiz, S.~Schulter, and M.~Chandraker, ``\BIBforeignlanguage{en-US}{Learning to simulate},'' \emph{\BIBforeignlanguage{en-US}{International Conference on Learning Representations,International Conference on Learning Representations}}, Sep 2018.

\bibitem{iso21448}
C.~ISO21448, ``Road vehicles—safety of the intended functionality,'' 2022.

\bibitem{survey}
W.~Ding, C.~Xu, M.~Arief, H.~Lin, B.~Li, and D.~Zhao, ``A survey on safety-critical driving scenario generation—a methodological perspective,'' \emph{IEEE Transactions on Intelligent Transportation Systems}, vol.~24, no.~7, pp. 6971--6988, 2023.

\bibitem{cat}
L.~Zhang, Z.~Peng, Q.~Li, and B.~Zhou, ``Cat: Closed-loop adversarial training for safe end-to-end driving,'' in \emph{Conference on Robot Learning}.\hskip 1em plus 0.5em minus 0.4em\relax PMLR, 2023, pp. 2357--2372.

\bibitem{survey2}
D.~Chen, M.~Zhu, H.~Yang, X.~Wang, and Y.~Wang, ``Data-driven traffic simulation: A comprehensive review,'' \emph{IEEE Transactions on Intelligent Vehicles}, 2024.

\bibitem{ral3}
J.~Fang, D.~Zhou, F.~Yan, T.~Zhao, F.~Zhang, Y.~Ma, L.~Wang, and R.~Yang, ``Augmented lidar simulator for autonomous driving,'' \emph{IEEE Robotics and Automation Letters}, vol.~5, no.~2, pp. 1931--1938, 2020.

\bibitem{ral4}
W.~Wang and D.~Zhao, ``Extracting traffic primitives directly from naturalistically logged data for self-driving applications,'' \emph{IEEE Robotics and Automation Letters}, vol.~3, no.~2, pp. 1223--1229, 2018.

\bibitem{advsim}
J.~Wang, A.~Pun, J.~Tu, S.~Manivasagam, A.~Sadat, S.~Casas, M.~Ren, and R.~Urtasun, ``Advsim: Generating safety-critical scenarios for self-driving vehicles,'' in \emph{Proceedings of the IEEE/CVF Conference on Computer Vision and Pattern Recognition}, 2021, pp. 9909--9918.

\bibitem{diffscene}
C.~Xu, D.~Zhao, A.~Sangiovanni-Vincentelli, and B.~Li, ``Diffscene: Diffusion-based safety-critical scenario generation for autonomous vehicles,'' in \emph{The Second Workshop on New Frontiers in Adversarial Machine Learning}, 2023.

\bibitem{art}
Q.~Zhang, S.~Hu, J.~Sun, Q.~A. Chen, and Z.~M. Mao, ``On adversarial robustness of trajectory prediction for autonomous vehicles,'' in \emph{Proceedings of the IEEE/CVF Conference on Computer Vision and Pattern Recognition}, 2022, pp. 15\,159--15\,168.

\bibitem{strive}
D.~Rempe, J.~Philion, L.~J. Guibas, S.~Fidler, and O.~Litany, ``Generating useful accident-prone driving scenarios via a learned traffic prior,'' in \emph{Proceedings of the IEEE/CVF Conference on Computer Vision and Pattern Recognition}, 2022, pp. 17\,305--17\,315.

\bibitem{vectornet}
J.~Gao, C.~Sun, H.~Zhao, Y.~Shen, D.~Anguelov, C.~Li, and C.~Schmid, ``Vectornet: Encoding hd maps and agent dynamics from vectorized representation,'' in \emph{Proceedings of the IEEE/CVF conference on computer vision and pattern recognition}, 2020, pp. 11\,525--11\,533.

\bibitem{vgg16}
K.~Simonyan and A.~Zisserman, ``Very deep convolutional networks for large-scale image recognition,'' \emph{arXiv preprint arXiv:1409.1556}, 2014.

\bibitem{densetnt}
J.~Gu, C.~Sun, and H.~Zhao, ``Densetnt: End-to-end trajectory prediction from dense goal sets,'' in \emph{Proceedings of the IEEE/CVF International Conference on Computer Vision}, 2021, pp. 15\,303--15\,312.

\bibitem{bbo}
P.~I. Frazier, ``A tutorial on bayesian optimization,'' \emph{arXiv preprint arXiv:1807.02811}, 2018.

\bibitem{bga}
A.~Lambora, K.~Gupta, and K.~Chopra, ``Genetic algorithm-a literature review,'' in \emph{2019 international conference on machine learning, big data, cloud and parallel computing (COMITCon)}.\hskip 1em plus 0.5em minus 0.4em\relax IEEE, 2019, pp. 380--384.

\bibitem{brs}
S.~Andrad{\'o}ttir, ``A review of random search methods,'' \emph{Handbook of simulation optimization}, pp. 277--292, 2014.

\end{thebibliography}



\end{document}